\documentclass[journal]{IEEEtran}
\usepackage{graphicx}
\usepackage{booktabs}
\usepackage{float}
\usepackage{subfig}

\usepackage[utf8]{inputenc}

\usepackage{blindtext}
 
\usepackage{array}

\begin{document}

\title{The Multi-Lane Capsule Network (MLCN)\thanks{\textbf{Acknowledgment}: This study was financed in part by CAPES/Brasil (Finance  Code  001), by CNPq (313012/2017-2), and by Fapesp (CCES 2013/08293-7).  This  work  was also supported  by  national funds  through  Fundação  para  a  Ciência  e  Tecnologia  (FCT) with reference FCT: UID/CEC/50021/2019.}}

\author{Vanderson M. do Rosario$^{1}$, Edson Borin$^{2}$, and Mauricio Breternitz Jr.$^{3}$
\thanks{$^{1}$ Ph.D. Candidate at IC, UNICAMP. Brazil. \textit{vanderson.rosario at ic.unicamp.br}}
\thanks{$^{2}$ Associate Professor at IC, UNICAMP. Brazil. \textit{edson at ic.unicamp.br}}
\thanks{$^{3}$ Invited Associate Professor at IST \& INESC-ID, Lisbon University. Portugal. \textit{mbreternitz at gmail.com}
}
}

\markboth{}%
{}

\maketitle

\begin{abstract}
We introduce Multi-Lane Capsule Networks (MLCN), which are a separable and resource efficient organization of Capsule Networks (CapsNet)~\cite{sabour2017dynamic} that allows parallel processing while achieving high accuracy at reduced cost. A MLCN is composed of a number of (distinct) parallel \emph{lanes}, each contributing to a dimension of the result,  trained using the routing-by-agreement organization of CapsNet. Our results indicate similar accuracy with a much-reduced cost in number of parameters for the Fashion-MNIST and Cifar10 datasets. They also indicate that the MLCN outperforms the original CapsNet when using a proposed novel configuration for the lanes. MLCN also has faster training and inference times, being more than two-fold faster than the original CapsNet in the same accelerator. 

%MLCN may be trained to reduce the overall number of lanes, and thus the HW cost to achieve a desired accuracy. Alternatively, MLCN may be trained for resiliency whereby the loss of a given lane minimizes the overall effect on the network operation. Finally, we indicate alternatives for HW specialization in which distinct lanes use different HW whereby a given target accuracy may be achieved with reduced resource costs.
%indicate faster training
%indicate potential for specializing each lane
%indicate ability to train for reduced number of lanes
% indicate ability to train for UNIFORM lanes adding to resiliency

\end{abstract}

% Note that keywords are not normally used for peer-review papers.
\begin{IEEEkeywords}
Capsule Network, Multi-lane, Deep Learning, CNN
\end{IEEEkeywords}

\IEEEpeerreviewmaketitle

% ----------------------------------------------------------------------------
\section{Introduction}
Deep Learning has become a widely used machine learning technique to solve many different problems from image processing to language translation to audio transcription amongst many others. In 2014, after the publication of the AlexNet architecture (stacking multiple layers of convolutions and maxpooling) \cite{krizhevsky2012imagenet}, deep learning became the state-of-art in image classification with the use of Convolutional Neural Networks (CNNs). One of the main mechanisms in these traditional CNNs are the Pooling operations that, although achieving some transitional invariance it loss a lot information. 
Subsequently, Sabour et al.~\cite{sabour2017dynamic}  proposed a novel approach to routing data in the network (dynamic routing algorithm) without using the traditional pooling mechanisms, demonstrated with a neural network called CapsNet. Towards this, it encodes features of the image as vectors and the dynamic routing algorithm is used to guarantee the global relationship between all of them. The traditional CNNs can 
miss such global relationships and, for instance, miss-classify an image that has a mouth, eyes, and nose as a face, independently of the order or relative position of these features. However, despite promising preliminary results, CapsNets are still a
young and not much-explored network. For example, one of the challenges that have been described using and testing CapsNet is that they have required larger training times.

Therefore, in this work we explore the CapsNet architecture, proposing a new organization for it. The original CapsNet is divided into multiple lanes that are data-independent and responsible to learn different dimensions of the vectors,  each lane learning distinct features. This organization outperforms the original CapsNet in training and inference time by adding more parallelism and reducing the number of trainable parameters. We also show that this organization also helps with the explainability of the network. All this, without losing learnability and generative performance. In fact, we show that one can construct a faster CapsNet divided into lanes that outperform the accuracy of the original one for the Fashion-MNIST \cite{xiao2017fashion} and Cifar10 \cite{krizhevsky2009learning} datasets.

% ----------------------------------------------------------------------------
\section{Related Work}

\subsection{CapsNet}

In 2011, Hinton, Krizhevsky, and Sida presented the idea of capsules \cite{hinton2011transforming}, a neural network wherein neurons have vectors as input and output, extending previously proposed scalars.  Later publication of the dynamic routing algorithm \cite{sabour2017dynamic}, which allows the dynamic choosing of the paths of activation of these capsules from one layer to another, greatly enhanced this organization. That work also presented an architecture that we will call the Capsule Network or CapsNet. 

Since the publication of the Dynamic Routing and the CapsNet, several works have emerged improving the algorithm or the architecture and experimenting with the power of CapsNet in other scenarios, applications, and datasets.
Shahroudnejad, Mohammadi, and Plataniotis \cite{shahroudnejad2018improved} presented an analysis of the explainability of CapsNet, showing that it has properties which help understand and explain its behavior. Jaiswal et al. 
\cite{jaiswal2018capsulegan} used the CapsNet in a Generative Adversarial Network (GAN) and showed that it can achieve lower error rates than simple CNN.
Ren and Lu  \cite{ren2018compositional} showed that CapsNet can be used for text classification and showed how to adapt the compositional coding mechanism to the CapsNet architecture. 
Jimenez-Sanchez, Albarqouni, and Mateus \cite{jimenez2018capsule} tested the CapsNet in Medical Imaging Data Challenges showing that it can achieve good performance even when having less trainable parameters than the tested counterpart CNNs.
Mobiny and Nguyen \cite{mobiny2018fast} tested the performance of CapsNet for lung cancer screening and showed that it could outperform CNNs mainly when the training set was small.
A similar result was achieved by Kim et al. in traffic speed prediction \cite{kim2018capsule} with CapsNet outperforming traditional CNNs approaches.
Mukhometzianov and Carrillo \cite{mukhometzianov2018capsnet} ran the CapsNet with multiple image datasets and they showed that although having good results, CapsNet still requires much more time to train than others CNNs.

While developing this work, Canqun et al. \cite{xiang2018ms} proposed the Multi-Scale CapsNet (MS-CapsNet). They proposed a division of the CapsNet network,  limited to three ``lanes" (they did neither name or explored the division concept), each with a different number of convolutions. In our work, we divide the CapsNet into more lanes, explore limitations of the division approach, the impact of dropout in feature division, the impact that such division has in training and inference speed and identify challenges. 
Another similar work, recently developed, was the Path Capsule Networks by Amer and Maul \cite{path2019} (Path-Capsnet) which explores the parallelism of CapsNets by splitting, with each path or lane being responsible for computing each digitcaps or primary capsule entirely, unlike the computation of different dimensions/features as in  MLCN. Thus, such approach does not help in explainability of the network and, although it does potentially provide a similar level of parallelism, the authors neither explored scalability or tested the approach in more complex datasets.

\section{Multi-lane Capsule Network}

\begin{figure}[!h]
    \center{
    \includegraphics[width=0.93\columnwidth]{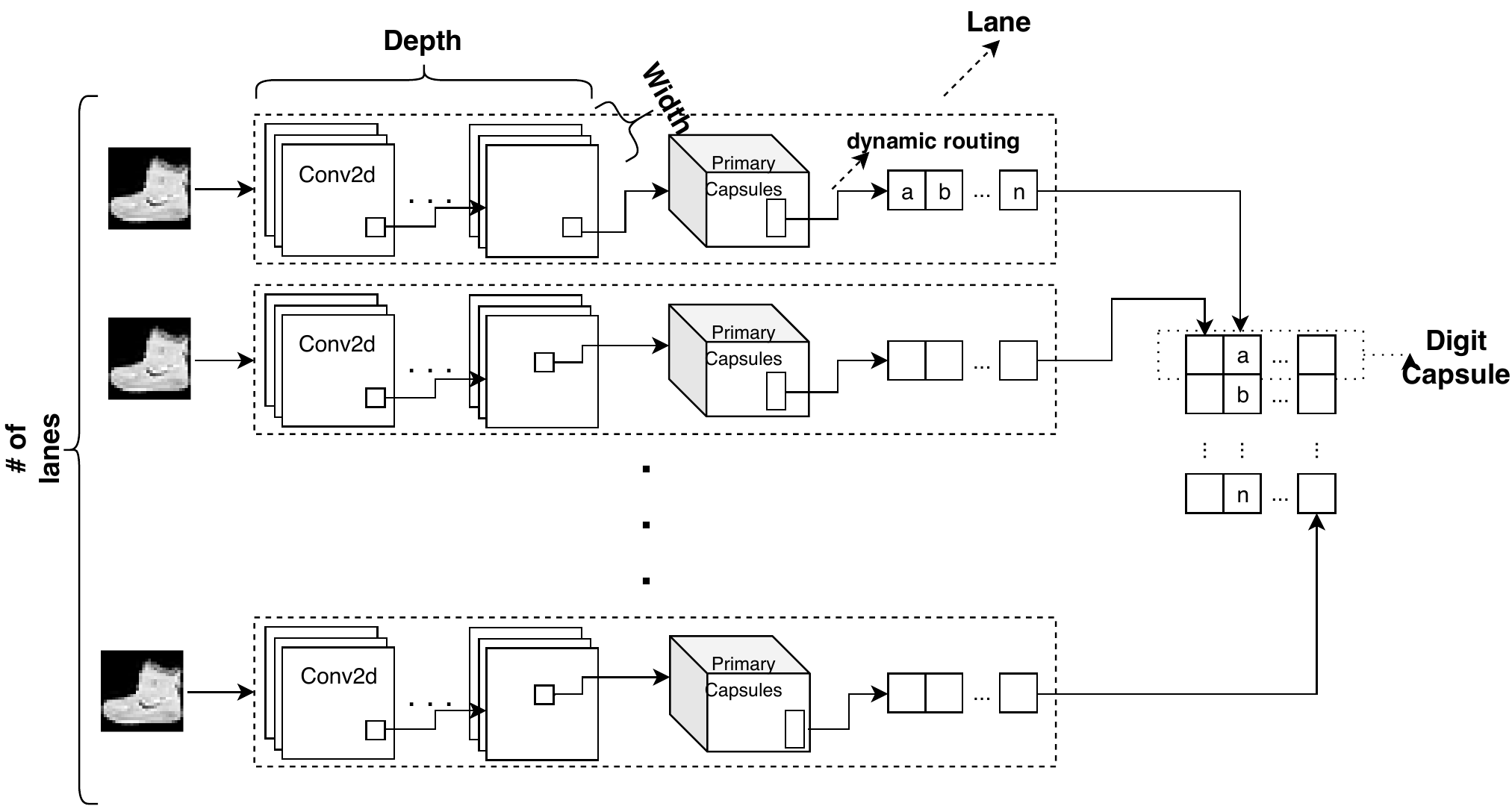}
    \caption{\label{fig:MLCNarch} Multi-lane Capsule Network Architecture}
    }
\end{figure}

The original version of CapsNet \cite{sabour2017dynamic} produces a set of $N$ Primary Capsules (PCs) by applying two convolutional steps to the original image. Each of these PCs, identified as $u_{i}$, is multiplied by a weight matrix $W_{i}$ and finally, a final set of capsules, the digit capsules, is created using the dynamic routing algorithm. Each of these digit capsule vectors represents one of the classes in the classification problem and the vector's length encodes the probability of the class being the one in the input image. However, more than just encoding the probability of a class, each vector also contains information to reconstruct the original image, with different dimensions of the vector representing different features of the image. Having this in mind, we propose to split the original CapsNet architecture \footnote{source code in https://github.com/vandersonmr/lanes-capsnet} (Figure \ref{fig:MLCNarch}), dividing the PCs into independent sets which we call lanes. Each of these sets of PCs, a lane, is responsible for one of the dimensions in the final digit capsules.

The number of PCs per lane may vary, as well as the way they are computed. In the original CapsNet, two 2D convolutions are applied to the input image and then reshaped to produce the PCs. More convolutions may be applied, what we call the $depth$ of a lane, or more filters can be used per convolution generating more capsules, what we call the $width$ of a lane. Further, distinct dimensions of a final digit capsule can be generated by lanes with different configurations (and thus distinct computational requirements).

There are two key advantages of this organization over the original CapsNet architecture. First, it allows parallelism of the execution, as each set of  PCs is constructed independently, improving performance and allowing training and deployment on distributed environments. Second, it improves the explainability of the network by associating different features of the image to each set of convolutions and PCs.

% ----------------------------------------------------------------------------
\section{Experimental Setup}

We tested our approach with the Fashion-MNIST and Cifar10 datasets.  Each experiment is performed three-times and we present the geometric mean.  There was no significant variance in the results, therefore we only present the means. Our experimental baseline refers to the original CaspNet configuration tested with the MNIST dataset \cite{sabour2017dynamic}. The baseline with MNIST creates 1152 PCs with dimension 8 and 10 digit capsules with dimension 16. We also use PCs with 8 dimensions and vary the number of PCs. We call a 1-size lane, a lane which creates $72$ PCs (same proportion of PCs per dimension as in the baseline) and a $k$-size lane a lane with $k\times72$ PCs. The experiment also varies the number of lanes (dimensions of the digit capsule), testing networks with 2, 4, 8, 16 and 32 lanes. All reported experiments were performed on a P100 NVIDIA GPU with 16gb of RAM and the training was performed with 20 epochs.

We tested two key variations of the lane configurations: the first,  mlcn1, has the same configurations as the baseline CapsNet and the second, mlcn2, includes differences that were found to increase significantly the performance of our architecture. The details of both are described below:

\begin{itemize}
    \item \textbf{Mlcn1:} in this configuration, each lane receives one copy of the original image, followed by one convolutional layer with $16 \times kernel\_size$ kernels, kernel size 9 and stride of 1. This is followed by another convolutional layer with $16 \times kernel\_size$ layers, kernel size 9 and strides of 2. The output of the second convolutions layer is reshaped into $72\times kernel\_size$ PCs with dimensions 8. Then dynamic routing is applied to generate one vector with dimension equal to the number of classes in the problem. In our tested datasets, there are 10 classes thus 10 dimensions.
    \item \textbf{Mlcn2:} in the second configuration, the lanes have first one 1x1 convolutional layer with $4 \times kernel\_size$ kernels, followed by two convolutional layers with kernel size 9 and stride of 1, but with $8 \times kernel\_size$ kernels, and one last convolution with $8 \times kernel\_size$ lanes, kernel size 9 and strides of 2. The output of these convolutions is then reshaped as in Mlcn1, outputting $72 \times kernel\_size$ PCs with dimension 8 and finally after the dynamic routing also producing one vector with a number of dimensions equal to the number of classes in the problem. 
\end{itemize}

\subsection{Dropout and Regularization in MLCN}

In both configurations, we notice that usually, a subset of lanes would actually be useful for the reconstruction and in classification. At a certain point, additional lanes stopped providing useful information, and new lanes just would produce similar results, adding no new information to the solution. To mitigate this effect we investigated a dropout approach. During the training process, we would discard 10\% of the lanes and only use the result of the other 90\% to calculate the final classification (for reconstruction all lanes were always used). This forces all lanes to contribute useful information to the solution of the problem. However, as we show and discuss in another section, this made the training process more laborious and caused all lanes to learn similar features. 

% ----------------------------------------------------------------------------
\section{Experimental Results}

\begin{figure}[!h]
    \center{
    \includegraphics[width=\columnwidth]{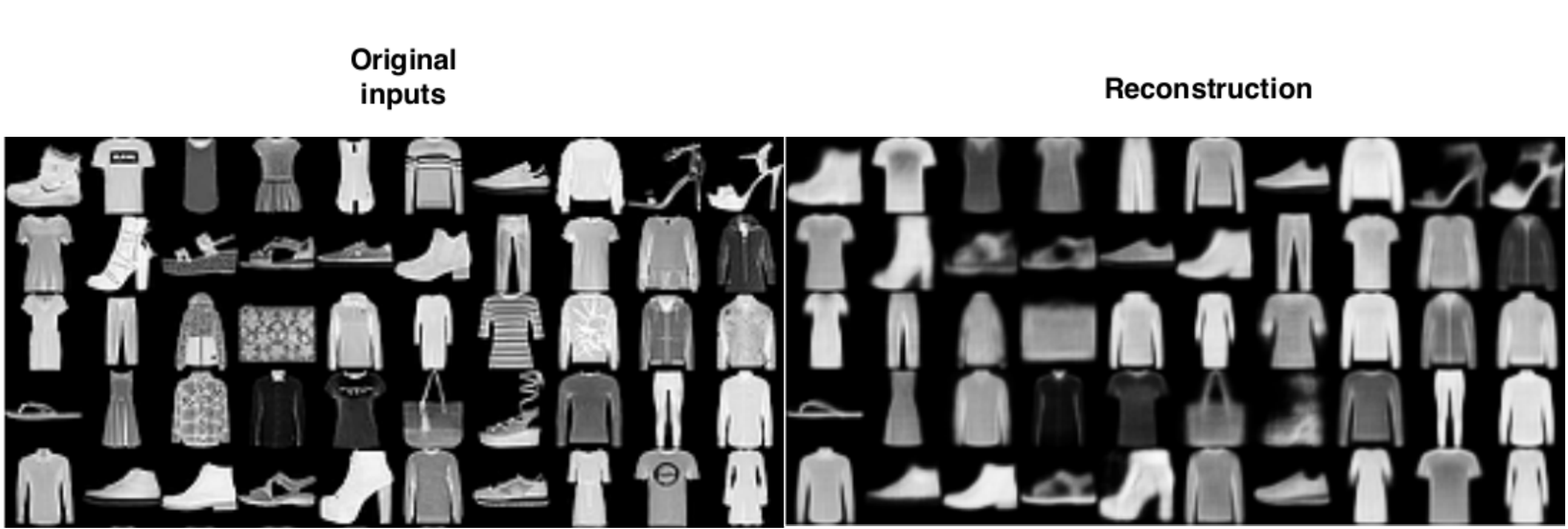}
    \caption{\label{fig:recon} Reconstructions from the fashion-mnist using MLCN.}
    }
\end{figure}

The digit capsule vectors which are constructed by concatenating the output of all lanes should have encoded information to entirely reconstruct the input image. During the training process, the longest digit capsule (its length encodes one class probability) is used as input to a fully connected neural network with two lanes of 512 and 1024 neurons, which produces a reconstruction of the original image. As seen in Figure \ref{fig:recon}, the output of 16 lanes (Mlcn1) can be used to reconstruct with high-fidelity the input image from the fashion-MNIST dataset. Therefore, it shows that dividing CapsNet into lanes does not prevent it from converging and learning image characteristics. 

\begin{figure}[!h]
    \center{
    \includegraphics[width=0.70\columnwidth]{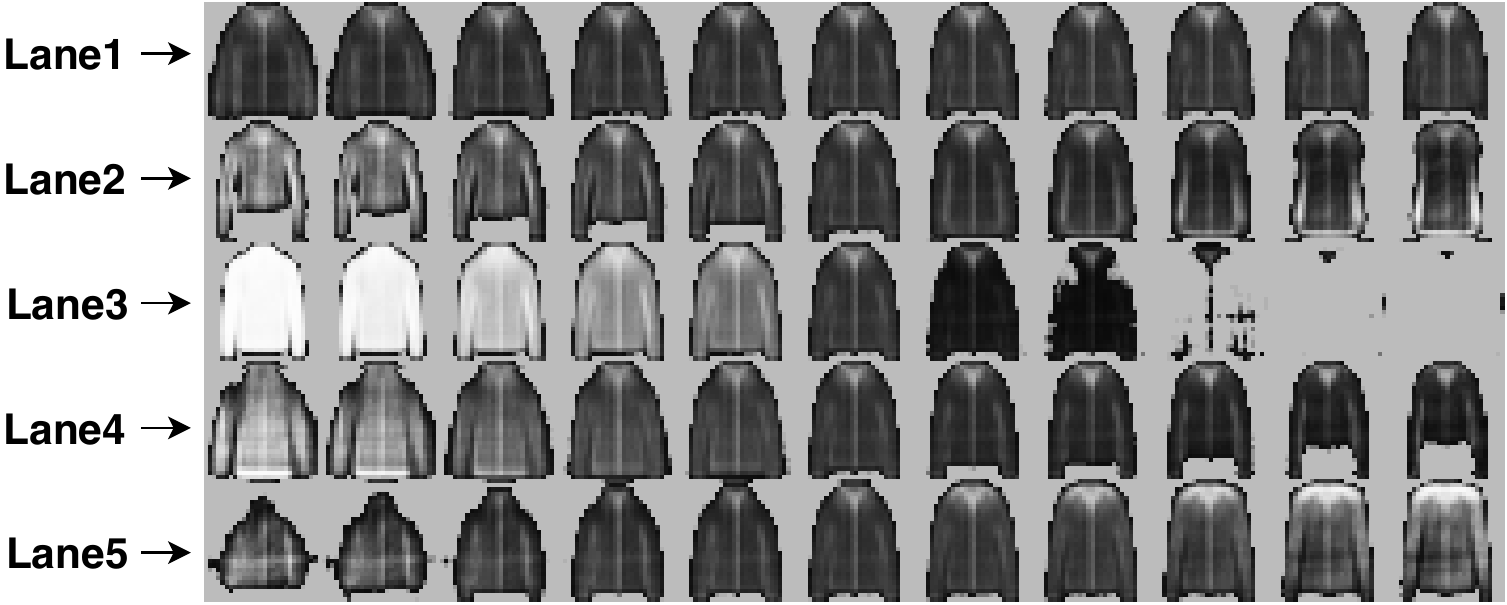}
    \caption{\label{fig:featurevariation} Synthetic variation on the lanes output.}
    }
\end{figure}

Furthermore, when we vary the output of each lane (adding -0.25, -0.2, -0.15, -0.1, -0.05, 0, 0.05, 0.1, 0.15, 0.2 and 0.25 to each dimension on the vector, one dimension at a time), we can observe the effect of this change on different features of the original image  (Figure \ref{fig:featurevariation}). 
Two key points from this: first, as each lane is entirely independent, we know that, for instance, lane5 and its associated convolutions are used to extract the size of the jacket. 
Further, we notice that similar properties were being extracted from different classes. 
So, for example, lane5 also extracts the size of the shoes. Second, we notice that adding more lanes than 5 for the MNIST dataset would result in lanes with no impact on the reconstruction and outputs essentially similar. 
Therefore, these lanes would not help to solve the classification problem. 
Solving this by forcing all lanes to extract different features of the image could allow the addition of more useful lanes. This is under exploration.

Although the second configuration, Mlcn2, improved the classification problem and increased the number of useful lanes to solve a specific problem (as we show in the next subsection), it did not have any visible impact on the reconstructed images, thus we only present images from Mlcn1.

\subsection{Number of Lanes}

\begin{figure}[!h]
    \center{
    \subfloat[Mlcn1 - fashion-MNIST]{\includegraphics[width=0.47\columnwidth]{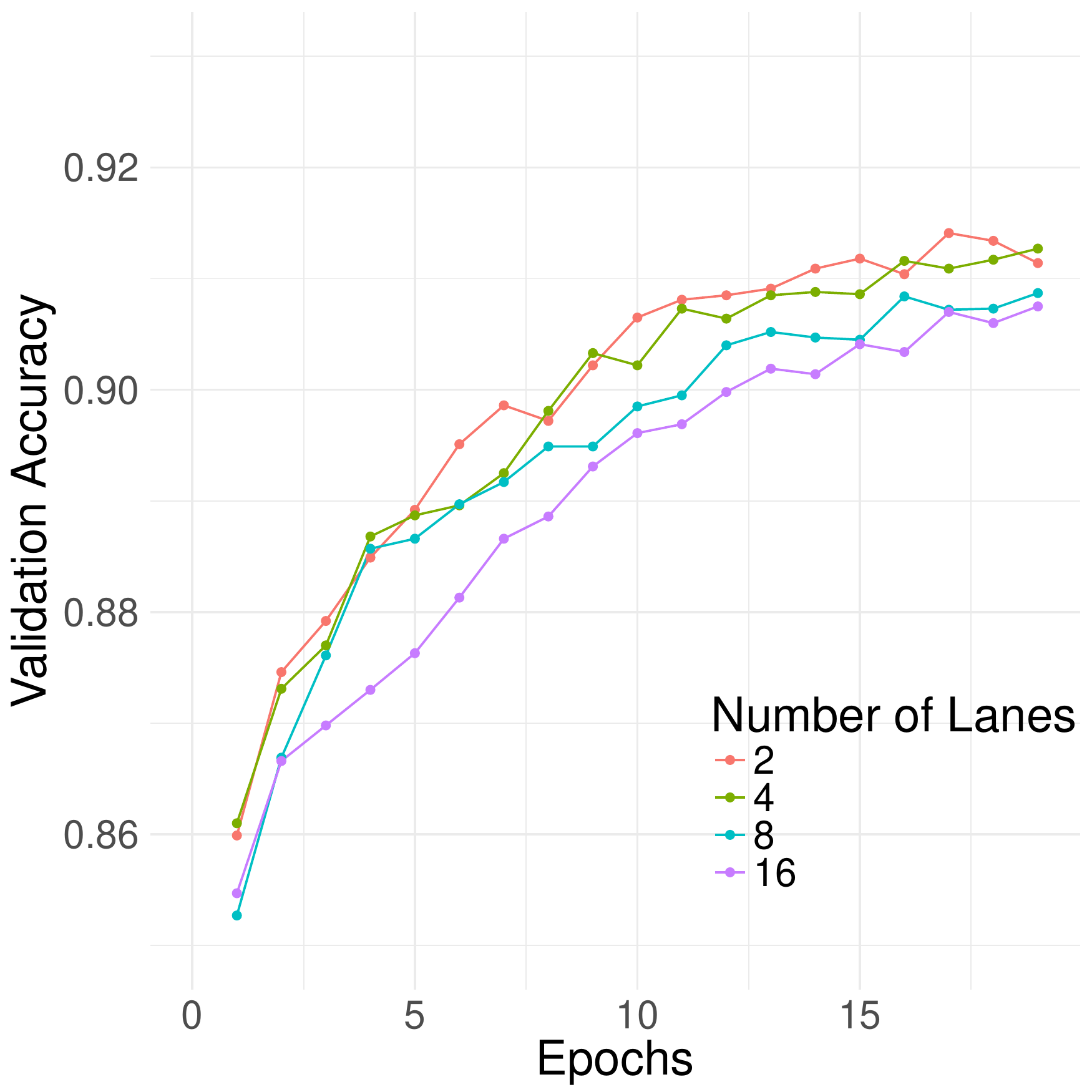}\label{fig:fashionlane1}}
    \subfloat[Mlcn2 - fashion-MNIST]{\includegraphics[width=0.47\columnwidth]{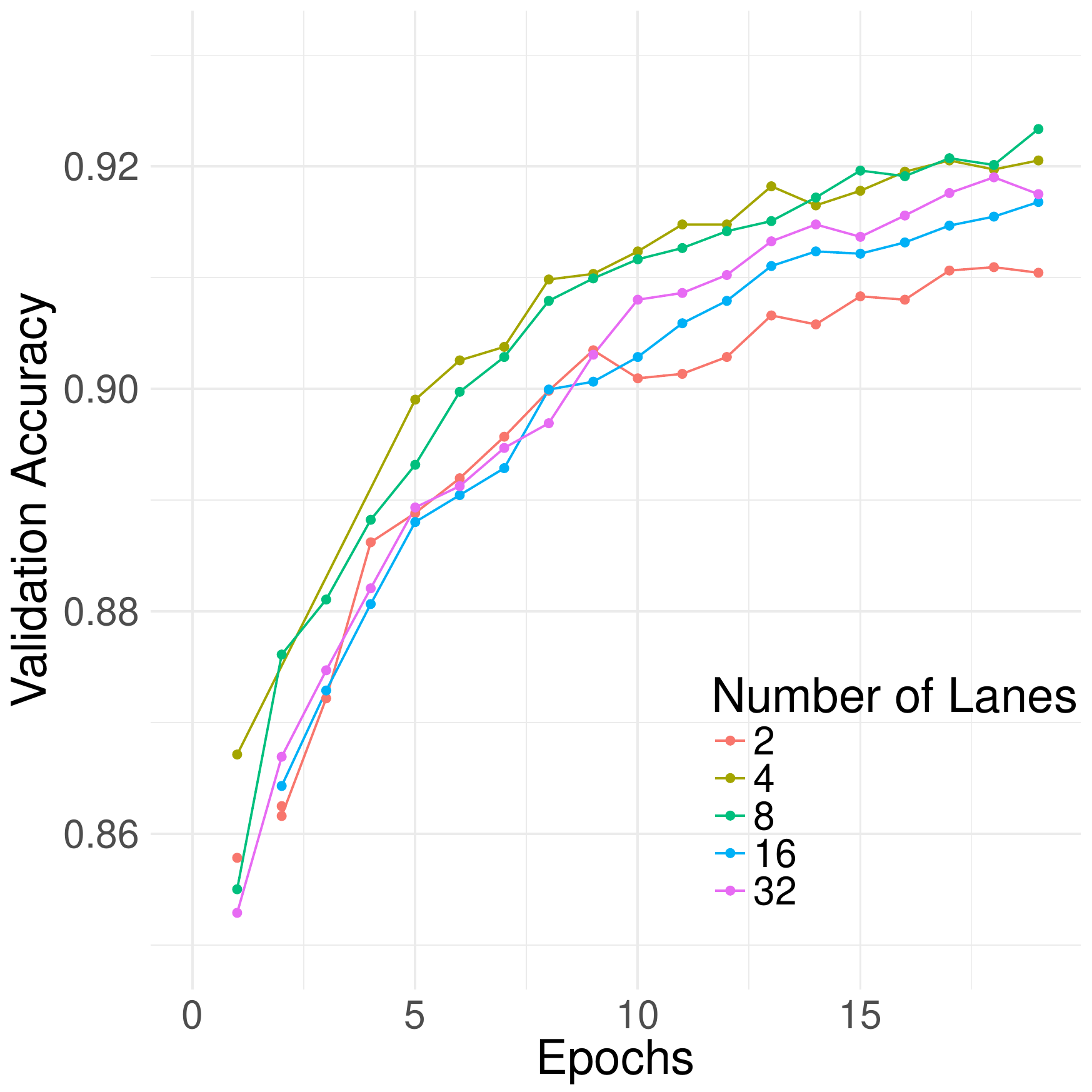}\label{fig:fashionlane2}}
    
    \subfloat[Mlcn1 - Cifar10]{\includegraphics[width=0.47\columnwidth]{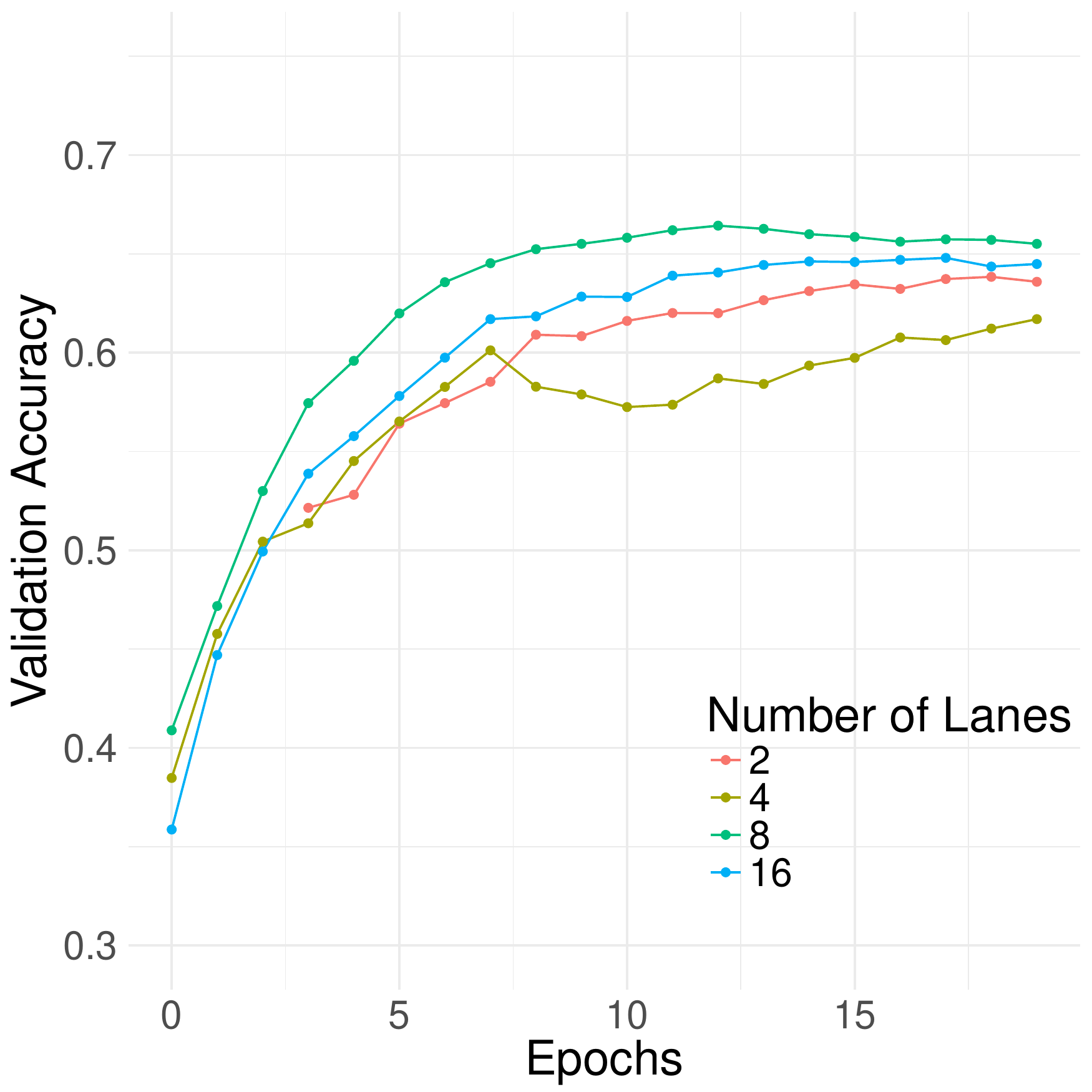}\label{fig:cifarlane1}}
    \subfloat[Mlcn2 - Cifar10]{\includegraphics[width=0.47\columnwidth]{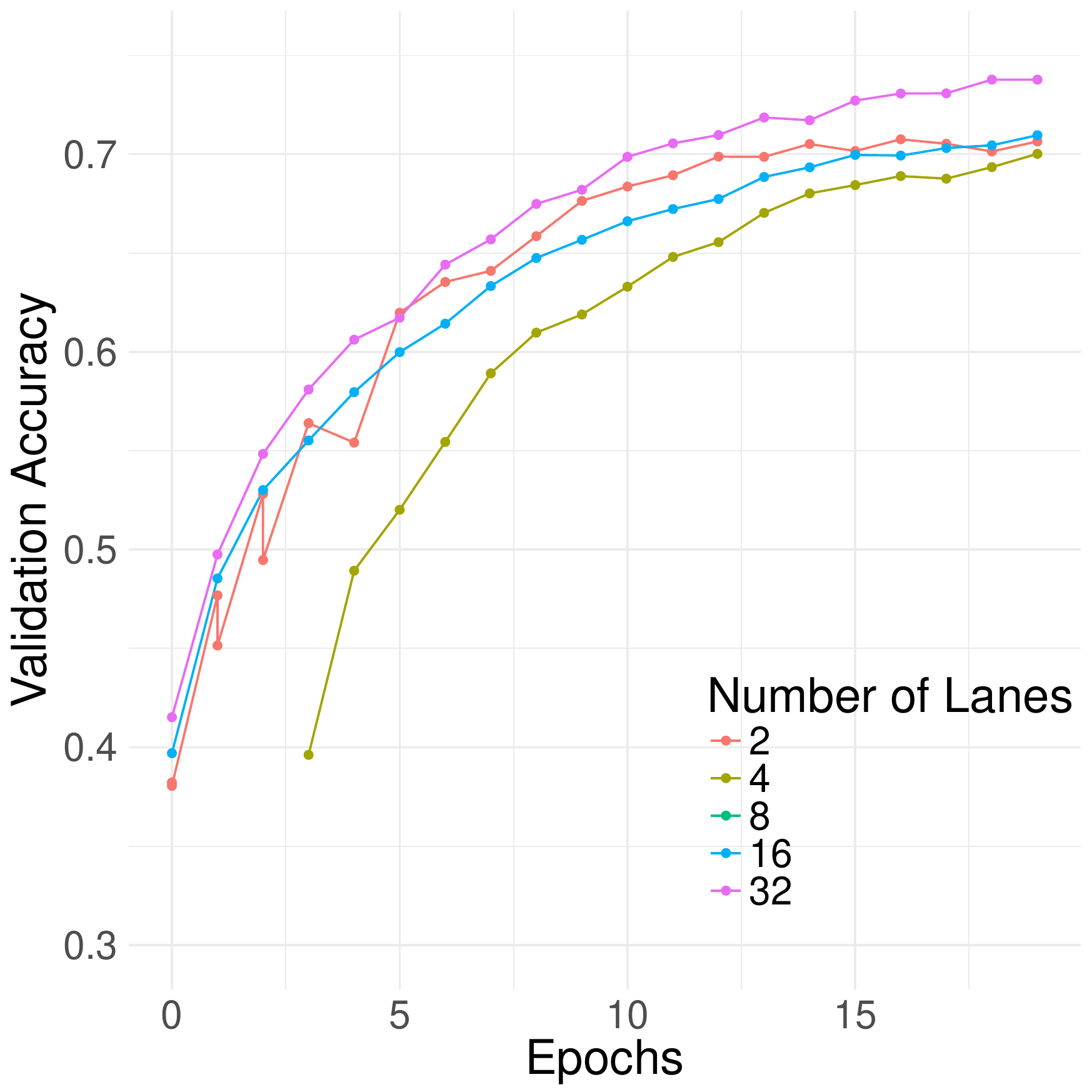}\label{fig:cifarlane2}}
    \caption{\label{fig:fashion} Varying number of lanes on Fashion-mnist and Cifar10}
    }
\end{figure}

As we noticed from the reconstructed images, adding more lanes does not improve the model, as new lanes stop learning new features. This same problem is reflected in the model accuracy. As seen in the classification results of  \ref{fig:fashionlane1} and \ref{fig:fashionlane2}, adding more lanes to  fashion-MNIST processing is not always beneficial. When using the first type of lanes (\ref{fig:fashionlane1}),  using only 2 or 4 lanes was better than using 8 or 16. First, because even adding more lanes, we would see rich features being learned in more than 4 lanes. Second, to effectively use near to 4 lanes, the learning rate of maximizing a vector with 4 dimensions is faster than when processing higher-dimensionality vectors. When using the second type of lane (\ref{fig:fashionlane2}) the scenario improves, with at least 8 dimensions being useful. However, it is still the case that using more lanes, 32 lanes, was not beneficial.

One thing to notice is that the number of lanes which will maximize the network performance is not only related to the lanes configurations, but also to the dataset. While fashion-MNIST appears to only need 4 lanes to represent its data with Mlcn1, Cifar10, a more complex dataset, requires 8 (\ref{fig:cifarlane1}) and with Mlcn2 it actually benefits from using 32 lanes (\ref{fig:cifarlane2}).  

Another important point from  Figures \ref{fig:fashionlane1}, \ref{fig:fashionlane2}, \ref{fig:cifarlane1} and \ref{fig:cifarlane2}  is that MLCN achieved better accuracy levels when using lanes of the second type.. This suggests that the second lane configuration is superior. When we compare the accuracy achieved by the two lanes types for both datasets with the baseline (Table \ref{tab:comp}), we see that Mlcn1 achieves similar accuracy rates as the baseline, but Mlcn is far better. Notice that when adding these characteristics from Mlcn2 into the baseline produces no performance gain.

\subsection{Lane's Size}

We also experiment with varying the size of lanes, increasing the number of convolution kernels and in this way adding more PCs. For all the results presented above, we used lanes of size 4 because it demonstrated for both datasets to be the better choice. We observed that increasing the lane's size increases the performance of the network, but for lanes larger than 4 it shows that this improvement starts to cease and the training rapidly overfits given the huge number of PCs.  

%\begin{figure}[!h]
%    \center{
%    \subfloat[fashion-mnist]{\includegraphics[width=0.51\columnwidth]{sizemnist.pdf}\label{fig:lanessize1}}
%    \subfloat[cifar10]{\includegraphics[width=0.51\columnwidth]{sizescifar.pdf}\label{fig:lanessize2}}
%    \caption{\label{fig:my-label}  Varying the lanes size on Fashion-mnist and Cifar10.}
%    }
%\end{figure}

\subsection{Dropout Trade-off}

To improve the number of lanes that learn useful features, we added a lane dropout mechanism (only for classification loss). Then, to test how individual lanes impact the classification, we removed one of the lanes from a $trained$ fashion-mnist MLCN with 16 Mlcn1 lanes and size 4. We then measure that lane's impact on the accuracy by calculating the new accuracy and getting the difference. Repeating this for all lanes results in a list of all lanes with individual accuracy impact. 

\begin{figure}[!h]
    \center{
    \subfloat[Accuracy]{\includegraphics[width=0.51\columnwidth]{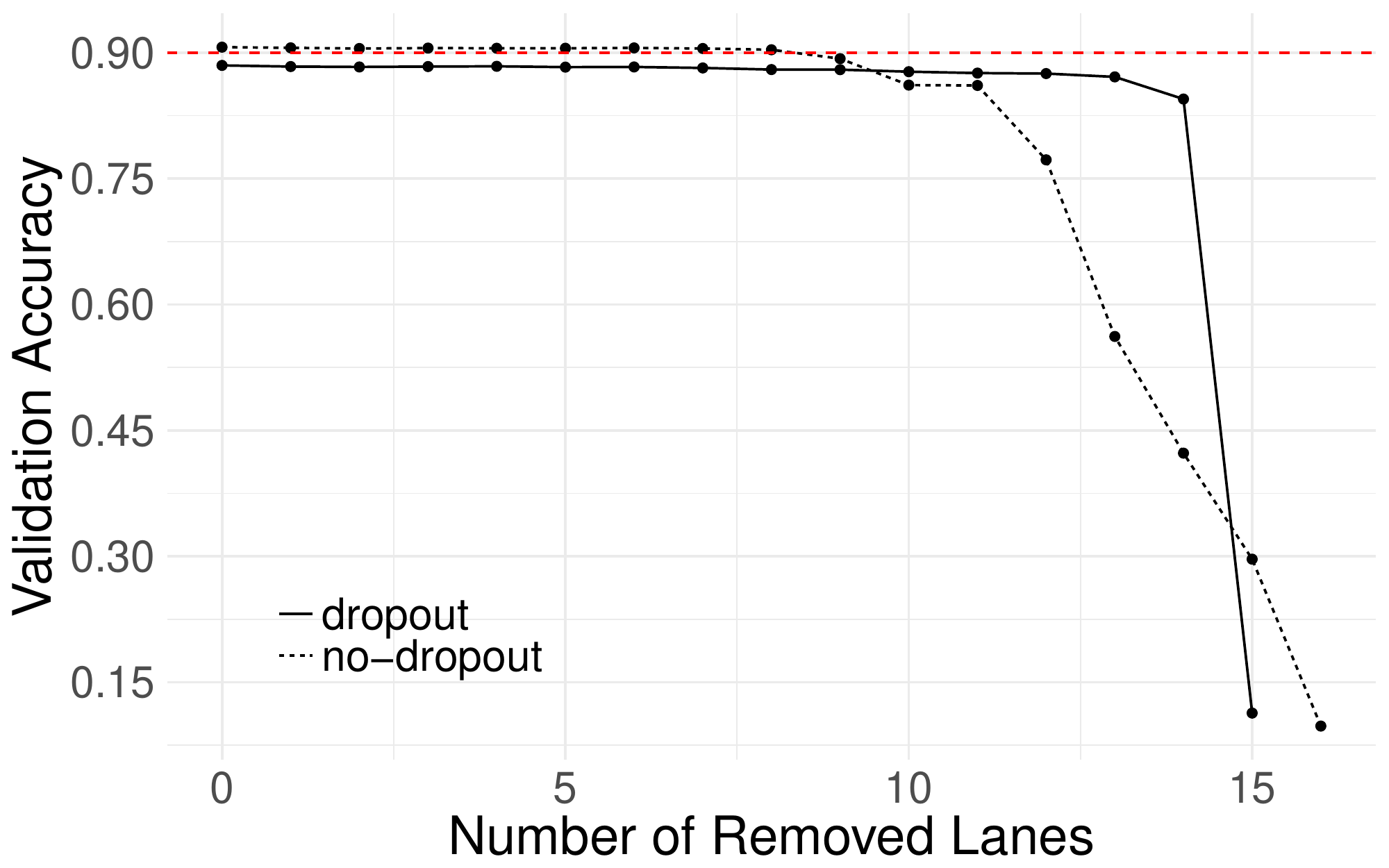}\label{fig:reduceacc}}
    \subfloat[Speed]{\includegraphics[width=0.51\columnwidth]{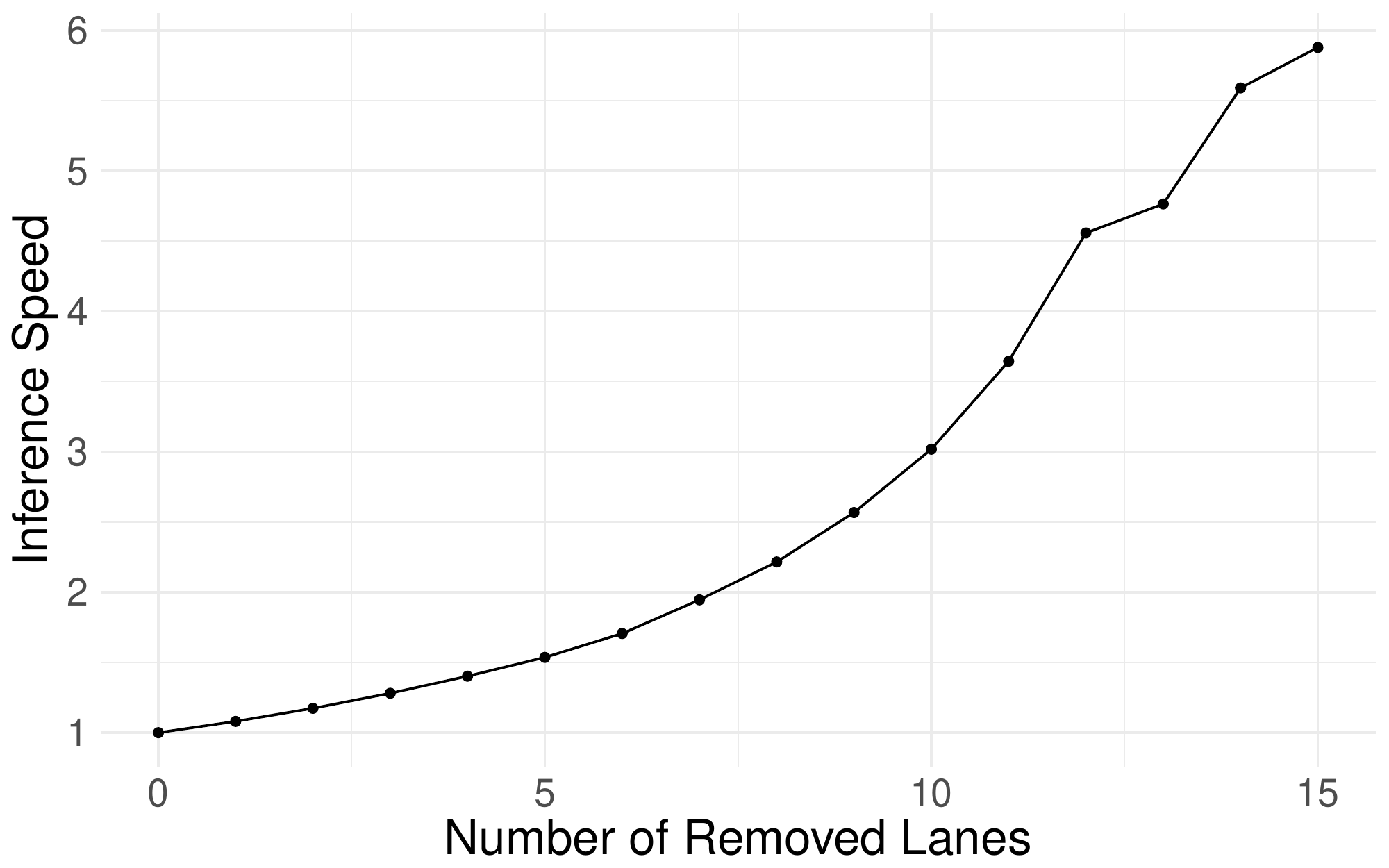}\label{fig:reducespeed}}
    \caption{\label{fig:my-label} Impact of removing lanes in order, starting with the less useful to the most.}
    }
\end{figure}

Figure \ref{fig:reduceacc} shows the impact of removing lane by lane sorted by the accuracy impact with and without dropout. Applying dropout generally reduced the maximum obtained accuracy. One of the reasons for this is that all lanes are thus forced to independently learn to classify the input, thus learning redundant information. This can be observed by the fact that when using the dropout we only have a significant reduction of accuracy after removing 15 lanes. In other words, 
many lanes are redundant for classification, so one can remove subsets and continue having a good result. However, without using dropout we have better maximum accuracy and, as seen in Figure \ref{fig:reduceacc}, there is less redundancy and more lanes contribute information for the classification, so their removal will severely impact accuracy.

Independently of using dropout, it is seen that, after the training process, one can reduce the size of the network by removing the least significant lanes without drastically impact the performance of the network. This not only impacts the size of the stored network but also its inference time and speed. Figure \ref{fig:reducespeed} shows how the inference speed (normalized with respect to the version with all lanes) increases as some lanes are removed. 

\subsection{MLCN Training and Inference Time}

To facilitate the comparison between Mlcn1 and Mlcn2 we chose their configurations to have the same amount of trainable parameters. Thus, both configurations had similar performance both in training and inference. Furthermore, when compared to the CapsNet,  MLCN needs a smaller or similar number of parameters to achieve the same accuracy. Added to the fact that the lanes are data-independent and have a higher amount of parallelism, we could train the MLCN much faster than the CapsNet. On average, for the same or better accuracy, one achieves a 2.4x speedup. Some of these results are shown in Table \ref{tab:comp}.

\begin{table}[!h]
\centering
\tiny
\caption{Comparison between Baseline and MLCN.}
\begin{tabular}[t]{l>{\raggedright}p{0.12\linewidth}p{0.05\linewidth}p{0.05\linewidth}p{0.1\linewidth}p{0.13\linewidth}p{0.1\linewidth}}
\toprule
Network & Set & \# of Lanes & Lane's Width & Params. & Train Time (seconds/epoch) & Accuracy\\
\midrule
Baseline & Cifar10 & - & - & 11,769,600 & 240 & 66.36\%\\
Mlcn1 & Cifar10 & 4 & 4 & 5,250,816 & 54 & 63.88\%\\
Mlcn1 & Cifar10 & 32 & 2 & 14,259,712 & 205 & 66.56\%\\
Mlcn2 & Cifar10 & 4 & 4 & 5,250,816 & 53 & 69.05\%\\
Mlcn2 & Cifar10 & 32 & 2 & 14,259,712 & 204 & \textbf{75.18}\%\\
\\
Baseline & Fashion-mnist & - & - & 8,227,088 & 220 & 91.30\%\\
Mlcn1 & Fashion-mnist& 2 & 4 & 3,655,376 & 21 & 91.14\%\\
Mlcn1 & Fashion-mnist& 8 & 4 & 10,633,232 & 90 & 90.87\%\\
Mlcn2 & Fashion-mnist & 2 & 4 & 3,655,376 & 20 & 91.01\%\\
Mlcn2 & Fashion-mnist & 8 & 4 & 10,633,232 & 92 & \textbf{92.63}\%\\
\bottomrule
\label{tab:comp}
\end{tabular}
\end{table}%
%TODO: add more lines into the table

\section{Conclusion}

In this work, we proposed a new organization for the CapsNet, introduced by Sabour et al. \cite{sabour2017dynamic}. We have shown that separating the CapsNet into lanes (MLCN) with each one responsible for one of the dimensions in the final capsules' vectors, and each one being data-independent, not only improves the training and inference time (being in average two time faster than the original CapsNet for similar numbers of parameters) but also improves the network final accuracy performance, outperforming the original CapsNet with reduced storage requirements.

As a next step, with these data-independent lanes, one can explore the training of very large lanes by using multiple systems or accelerators in a distributed scheme. Addressing the challenge of scaling  MLCN in number of lanes, finding novel training mechanisms,  and benchmarking against competitive approaches such as MS-CapsNet and Path-CapsNet are the subject of future work.

\newpage
\bibliographystyle{unsrt}
\bibliography{ref}

\end{document}